\documentclass[conference]{IEEEtran}
\IEEEoverridecommandlockouts
\usepackage{cite}
\usepackage{amsmath,amssymb,amsfonts}
\usepackage{graphicx}
\usepackage{textcomp}
\usepackage{xcolor}
\usepackage{lscape}
\usepackage{booktabs}
\usepackage{wrapfig}
\usepackage{algorithm}
\usepackage[noend]{algpseudocode}
\usepackage[utf8]{inputenc}

\def\BibTeX{{\rm B\kern-.05em{\sc i\kern-.025em b}\kern-.08em
    T\kern-.1667em\lower.7ex\hbox{E}\kern-.125emX}}
\usepackage{caption}
\begin{document}

\title{You Only Look Twice! for Failure Causes Identification of Drill Bits
}

\author{\IEEEauthorblockN{Asma Yamani\IEEEauthorrefmark{1},
Nehal Al-Otaiby\IEEEauthorrefmark{1},
Haifa Al-Shemmeri\IEEEauthorrefmark{1},
and Imane Boudellioua\IEEEauthorrefmark{1}
} \\
\IEEEauthorblockA{\IEEEauthorrefmark{1}Information and Computer Science Department, KFUPM \\ \{g201906630,g202110590, g202110890,imane.boudellioua\}@kfupm.edu.sa \\
}
}
\maketitle

\begin{abstract}
Efficient identification of the root causes of drill bit failure is crucial due to potential impacts such as operational losses, safety threats, and delays. Early recognition of these failures enables proactive maintenance, reducing risks and financial losses associated with unforeseen breakdowns and prolonged downtime. Thus, our study investigates various causes of drill bit failure using images of different blades. The process involves annotating cutters with their respective locations and damage types, followed by the development of two YOLO Location and Damage Cutter Detection models, as well as multi-class multi-label Decision Tree and Random Forests models to identify the causes of failure by assessing the cutters' location and damage type. Additionally, RRFCI is proposed for the classification of failure causes. Notably, the cutter location detection model achieved a high score of 0.97 mPA, and the cutter damage detection model yielded a 0.49 mPA. The rule-based approach overperformed both DT and RF in failure cause identification, achieving a macro-average F1-score of 0.94 across all damage causes. The integration of the complete automated pipeline successfully identified 100\% of the 24 failure causes when tested on independent sets of ten drill bits, showcasing its potential to efficiently assist experts in identifying the root causes of drill bit damages.

\end{abstract}

\section{Introduction}
\label{sec:sample1}
Drill bits are mechanical machinery commonly used during the oil and gas discovery and extraction phases and are usually composed of PDC cutters~\cite{ashok2021drill}. The International Association of Drilling Contractors (IADC) proposed a new classification system for diamond drill bits cutters based on their characteristics in 1987~\cite{winters19871987}. Several types were documented, including bits with natural diamond and polycrystalline diamond compact (PDC). 
Damage to PDC cutters can occur due to both normal and dysfunctional drilling. These dysfunctions may originate from various down-hole conditions\cite{witt2021quantifying}. The main goal of IADC dull grading was to capture the bit's operational damage and help petroleum companies enhance drilling performance. However, because the grading process is manual, the bit-grading output would be considered an inconsistent measurement. Thus, the drill bit investigation team currently requires experts to examine the damaged bit, find the cause, and then assess it according to their findings. This process is costly and time-consuming~\cite{ashok2021drill}. Identifying the root cause of failure efficiently aids in preparation for future bit runs and prolongs the life expectancy of the drill bit.\par

 Recent advances in deep learning have enabled consistent and automatic bit grading~\cite{ashok2020drill}. From a computer vision perspective, identifying the cutters of the drill bit to evaluate its condition is a small object detection task due to the fact that these cutters typically fill less than 20\% of an image; as per one of the definitions of small objects~\cite{zhu2016traffic}. Small object detection is a challenging problem as it is greatly affected by several factors including image's resolution and the varying architectural representation of the detected parts in the learning models~\cite{liu2021survey}. Another limitation lies in the challenging and time-consuming data annotation process, as multiple small objects in the same images need to be annotated by experts\cite{shorten2019survey}. In this paper, we describe our integrative approach for automatic identification of drill bits damage causes to answer the following research questions: \textbf{How to identify the root cause of drill bit failure using bit images?} 
 
The rest of the paper is organized as follows: Section \ref{sec:sample1} provides an overview of PDC cutters and their prospective failures and relevant machine learning algorithms. Section \ref{lr}, provides an overview of previous work. Section \ref{meth}, presents the methodology and our research procedure. Results and analysis are in section \ref{res}. In section \ref{des}, a detailed discussion of the obtained results and limitations. Finally, the conclusion and future works are in section \ref{conc}.

\section{Background}

\label{sec:sample1}
\subsection{PDC Damage and Root Causes}
\begin{figure}[htb]
  \centering
  \includegraphics[width=0.25\columnwidth]{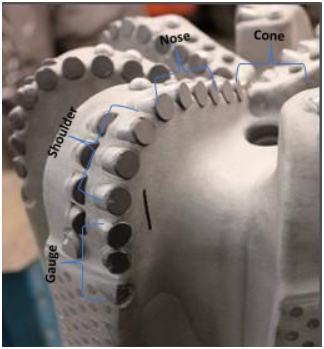}
  \caption{PDC bit structure.}
  \label{fig1}  
\end{figure}
\begin{figure}[!ht]
  \centering
  \includegraphics[width=0.7\columnwidth]{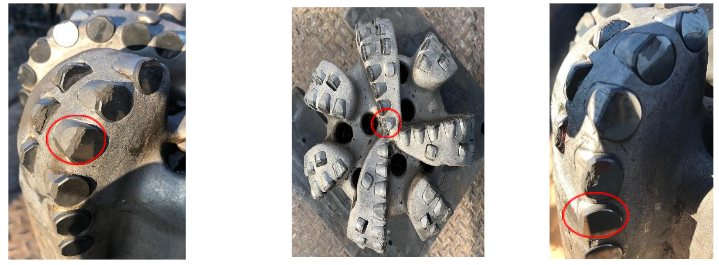}
  \caption{Smooth Wear in the Shoulder (left), in the Cone (middle), in the Gauge (right).}
  \label{fig2}
\end{figure}

The most common failures of the PDC cutters, which are covered in this work, are (Ringout (Nose or Shoulder), Core out, Smooth Wear, Fracture (Normal or Tangential), and Thermal Damage)~\cite{witt2021quantifying}. Damage to PDC cutters can occur due to both normal and dysfunctional drilling~\cite{ashok2021drill}. These dysfunctions occur due to various down-hole conditions, including Whirl, stick-slips, and Axial~\cite{witt2021quantifying}. PDC failures can be detected using its visual representation across different spots in the drill bit, such as Nose, Shoulder, Cone, and Gauge. Fig.~\ref{fig1} shows a typical drill bit image where failure can be inferred from thermal, mechanical, and chemical origins~\cite{witt2021quantifying,witt2021methodology}. Whirl damage is most common in the Gauge, Shoulder, or Nose compared to the Cone ~\cite{witt2021quantifying,ashok2021drill}. Deep drilling inside the hole by friction torque between the bit and the rock, followed by unexpected release, is referred to as stick-slip. The latter affects the bits in the Gauge, Shoulder, Nose, and Cone~\cite{ashok2021drill}. In another scenario, damage can occur during the stick stage where the load on the PDC cutters is vertical due to the bit's stability. This type of 
damage is referred to as 'Axial load'~\cite{witt2021methodology}. Additionally, this type of damage occurs more frequently in the Nose and the Shoulder~\cite{ashok2021drill}. One of the simplest forms of damage is Smooth Wear, characterized by its straight wear, as shown in Fig.~\ref{fig2}.  can generally be found at either the Shoulder or the Cone. However, it tends to appear frequently in the Shoulder more than in the Cone, and this is due to smooth cutters going in a greater distance each rotation than Cone cutters~\cite{witt2021methodology}. 
And from our observation, this kind of failure also appears in the Gauge.

\begin{figure}[!ht]
  \centering
  \includegraphics[width=0.5\columnwidth]{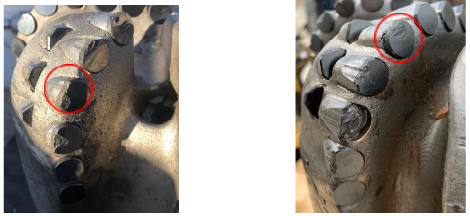}
  \caption{Normal Fracture in the Shoulder (left), in the Nose (right).}
  \label{fig3}
\end{figure}

\begin{figure}[!ht]
  \centering
  \includegraphics[width=0.7\columnwidth]{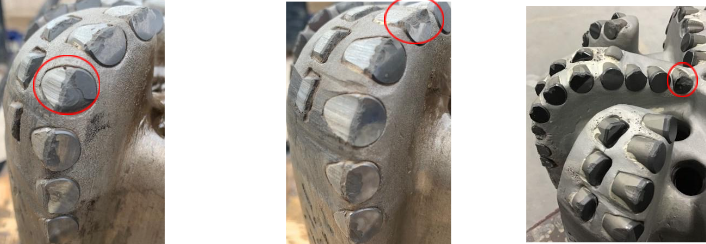}
  \caption{Tangential Fracture in the Shoulder (left), in the Nose (middle), and in the Cone (right).}
  \label{fig4}
\end{figure}

Fracture damage can be classified according to the cutter loading direction. Either the load will be in the direction of the rotation (tangential) or along the cutter edge (normal). Normal Fracture is caused by mechanical loading along the cutter's face~\cite{witt2021methodology}. Moreover, this can appear on the Shoulder due to rock hardness (Whirl) pressing on the Shoulder, in the Nose caused by Axial, or even in the Cone due to sliding distance (stick-slip), as illustrated in Fig.~\ref{fig3}~\cite{pastusek2018drilling}.
The Tangential Fracture is caused by high front face loading opposite to the direction of the cutter's movement. It can be visually characterized by a crack along the cutter, which does not extend beyond the cutter surface~\cite{witt2021quantifying}. In addition, it can be seen in the Nose caused by Whirl, or in the Cone due to sliding distance (stick-slip) facing the Cone, or in the Shoulder due to rock hardness (Whirl) as shown in Fig.~\ref{fig4}~\cite{pastusek2018drilling}.

\par \begin{wrapfigure}{l}{0.25\columnwidth}

  \centering
  \includegraphics[width=0.25\columnwidth]{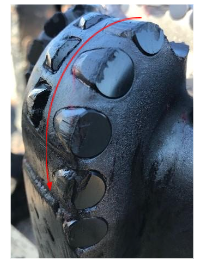}
  \caption{Thermal Damage.}
  \label{fig5}
\end{wrapfigure}
\begin{figure}[!ht]
  \centering
  \includegraphics[width=0.54\columnwidth]{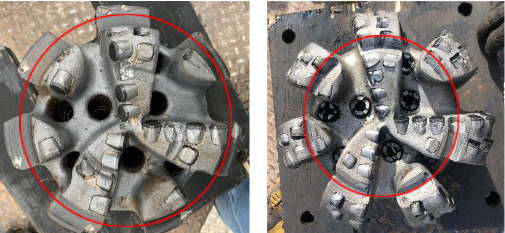}
   \includegraphics[width=0.25\columnwidth]{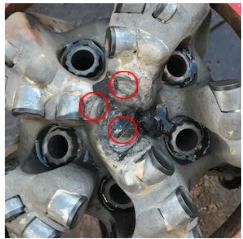}
  \caption{Ringout in the Shoulder (left), in the Nose (middle), Core-out (right).}
  \label{fig6}
\end{figure}

A thermal origin can be another cause of the PDC cutter damage caused by the interstitial cobalt that persists between the diamond grains. It appears around the margin of a worn scar and is surrounded by equally damaged cutters, as shown in Fig.~\ref{fig5}~\cite{witt2021quantifying}. If one cutter is destroyed as the bit rotates, the cutter in the next radial position does additional work. Thus, all cutters at the same radial position start to Ringout or core out, in Fig.~\ref{fig6}~\cite{witt2021methodology}. Ringout represents one of the most frequent damages that can occur in the drill bit due to systematic degradation of the cutter structure caused by hard rock drilling (Whirl)~\cite{witt2021methodology,pastusek2018drilling}. Usually, this kind of damage occurs in the Shoulder and the Nose, whereas core-out damages correspond to the failed Cones~\cite{witt2021methodology}.
 

\subsection{Machine Learning Approaches}

One widely used machine learning classification model is the Decision Tree (DT). A DT classifies the data in a tree-structure format to observe valuable patterns in diverse datasets. The resulting tree would be used to make sophisticated decisions\cite{breiman2001random,priyanka2020decision}.
 However, DT models have limitations, including scalability, height optimization, and instability. To overcome that, Random Forest (RF) was proposed~\cite{breiman2001random}, which works with multiple trees by combining tree predictors based on random vector values that are partitioned independently. The distribution of this vector is the same for all trees in the forest. Thus, the number of independent trees in a forest and their association affect the classification error. RF provides many robust characteristics, such as effectively eliminating anomalies and noise~\cite{breiman2001random}.

Recent advances in deep learning have enabled the adoption of pre-trained models for one-stage object detectors. An example of such an approach is YOLO~\cite{redmon2016you,redmon2017YOLO9000,redmon2018YOLOv3}. YOLO is a deep learning method that can be utilized to automate inspection for object detection at a high detection rate and speed ~\cite{ultralyt31:online}. 
YOLOv5 is built using a CNN with Darknet architecture as a backbone. It splits the image into grids, each having an equal dimensional region of SxS. For each cell, it performs the detection and classification at once for B boxes. It is significantly lighter than previous models, which makes it more suitable for real-time detection~\cite{yan2021real}. 
\section{Related Work}
\label{lr}

An approach was proposed in~\cite{ashok2020drill} to classify the damage on drill bits. The first step was to use a pre-trained YOLOv3 model to recognize all the cutters on a drill bit image using Convolutional Neural Networks (CNN). Image processing techniques were used to calculate the damage to each edge. The damage evaluation of the drill bit was done using the information acquired in the preceding steps. The YOLOv3 model accurately recognized the cutters in an image. The accuracy of cutter identification was further enhanced by heuristics that forecasted likely cutter positions based on blade location and shape. However, distinguishing distinct cutters from a group of images of the same bit was challenging since the photographs could not be stitched together correctly. Consequently, each image is independently rated, and these individual assessments are amalgamated to form a comprehensive evaluation of the bit. Furthermore, not all detected cutters can be categorized as damaged or intact. In summary, this study represents a crucial initial step in automating the assessment of bit damage and extracting valuable information from used drill bits. Applying deep learning techniques to cutter identification, damage detection, and quantification can significantly enhance bit design, selection, and drilling efficiency. Following, the authors of a previous study~\cite{ashok2020drill} extended their work in~\cite{ashok2021drill} to provide a software algorithm that can automatically evaluate 2D-bit images and determine the cause of bit damage and failure. They correlated the damage visible in the bit images with a dataset's root cause of failure and created a classifier to determine the underlying cause immediately. Their dataset only included a few types of failures and bit damage, so it was supplemented with bit pictures for which a subject-matter expert's examination identified the kind of failure. One significant finding was that bit pictures are not permanently recorded correctly, which affects the method's accuracy. The algorithm also allows for process standardization and uniformity in bit analysis throughout a company's activities. Image processing, drilling settings, and bit selection may also be adjusted to extend future bit life by detecting the likely underlying causes of PDC damage. \par Moreover, the authors of~\cite{witt2021quantifying} have introduced a straightforward method to predict the wear of PDC bits. The technique helps identify if there is any failure in the bit during drilling operations. Also, it measures wear and failure using real-time surface sensor data and PDC dull pictures. Based on the research, it was discovered that when drilling complex formations with PDC bits, if the bit is Damaged Beyond Repair (DBR), such as being rung out, subsequent bit runs in the same well performed worse than in cases where the previous bit was pulled before substantial damage occurred. Pulling and replacing bits before they reach the DBR condition can improve overall drilling performance. The proposed wear metric is easy and inexpensive to apply, making it essential for low-cost land wells, and it requires real-time surface sensor data. This approach allows for a more focused evaluation of PDC bit wear, enhances drilling performance, and determines appropriate bit pull criteria. A recent study~\cite{watson2022iadc} introduced a forensic workflow that connects drilling dysfunction to drill bits and Bottom Hole Assembly (BHA) damages. The study offers a hierarchical model that explains how data can be analyzed in different domains, including time, depth, and frequency, which can be helpful in forensic evaluation. Moreover, the study provides a summary of the various drilling mechanic calculations that are commonly used for identifying dysfunction. The procedure outlined in the study has been modified to facilitate data gathering for forensic analysis and serves as a guide for drilling experts in the field and office. 

\section{Materials and Methods} 
\label{meth}

\subsection{Procedure.} 
\begin{figure}[h]
  \centering
  \includegraphics[width=\columnwidth]{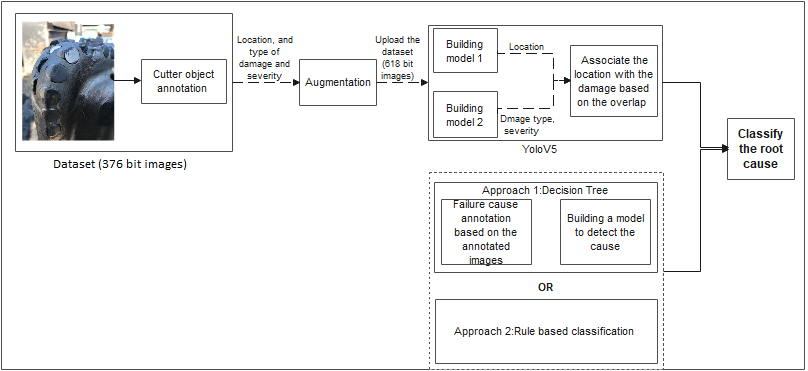}
  \caption{Methodology.}
  \label{fig9}
\end{figure}

The pipeline to identify the bit failure cause consists of several steps, as illustrated in Fig.~\ref{fig9}. The dataset used in the pipeline consists of 376-bit images with different views of the drill bits. These images are annotated, preprocessed, and augmented for the respective models. Four machine learning models are then built; two are object detection models using YOLOv5 based on the location and damage type. The third and fourth models are designed to detect the bit failure cause based on the detected damages per location. The pipeline also has a complementary rule-based approach for detecting the cause of bit failure.

\vspace{-\baselineskip}
\subsection{Data.}

\subsubsection{Dataset Description}
We utilized the dataset from~\cite{WittDoerring2021}. Most of the image sets contained the top view of the drill bit and seven images for the side views of the different main blades of the cutter. Each blade contained six to ten cutters. A total of 46 image sets were utilized from the dataset, bringing the total images to 376, corresponding to 46 drill bits.

\subsubsection{Dataset Annotation}
We annotated the dataset with respect to the following characteristics: cutter location, cutter type of damage, and failure cause. Each of the annotation phases are described below.

\begin{figure}[!ht]
  \centering
  \includegraphics[width=0.35\columnwidth]{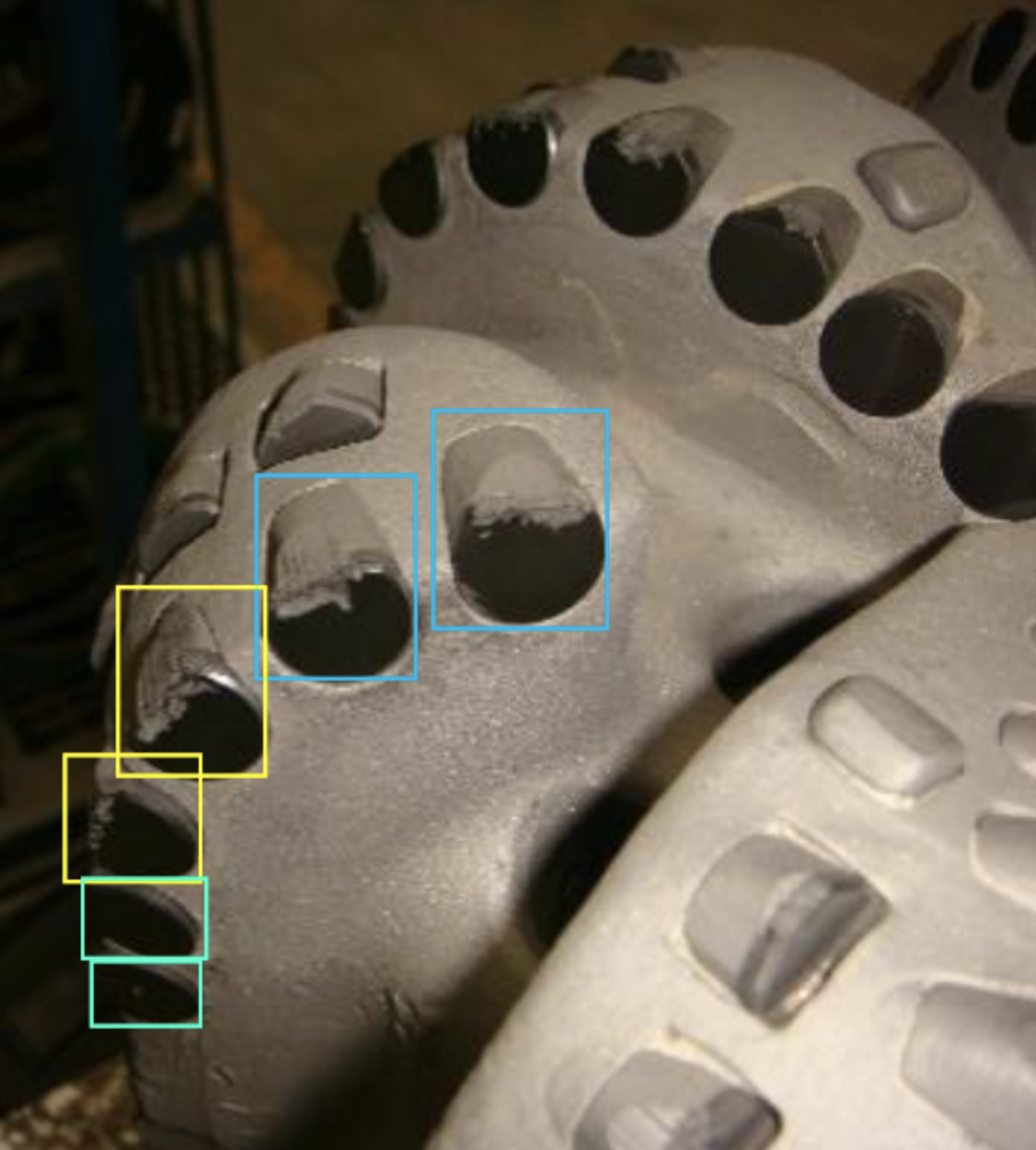}
  \includegraphics[width=0.3\columnwidth]{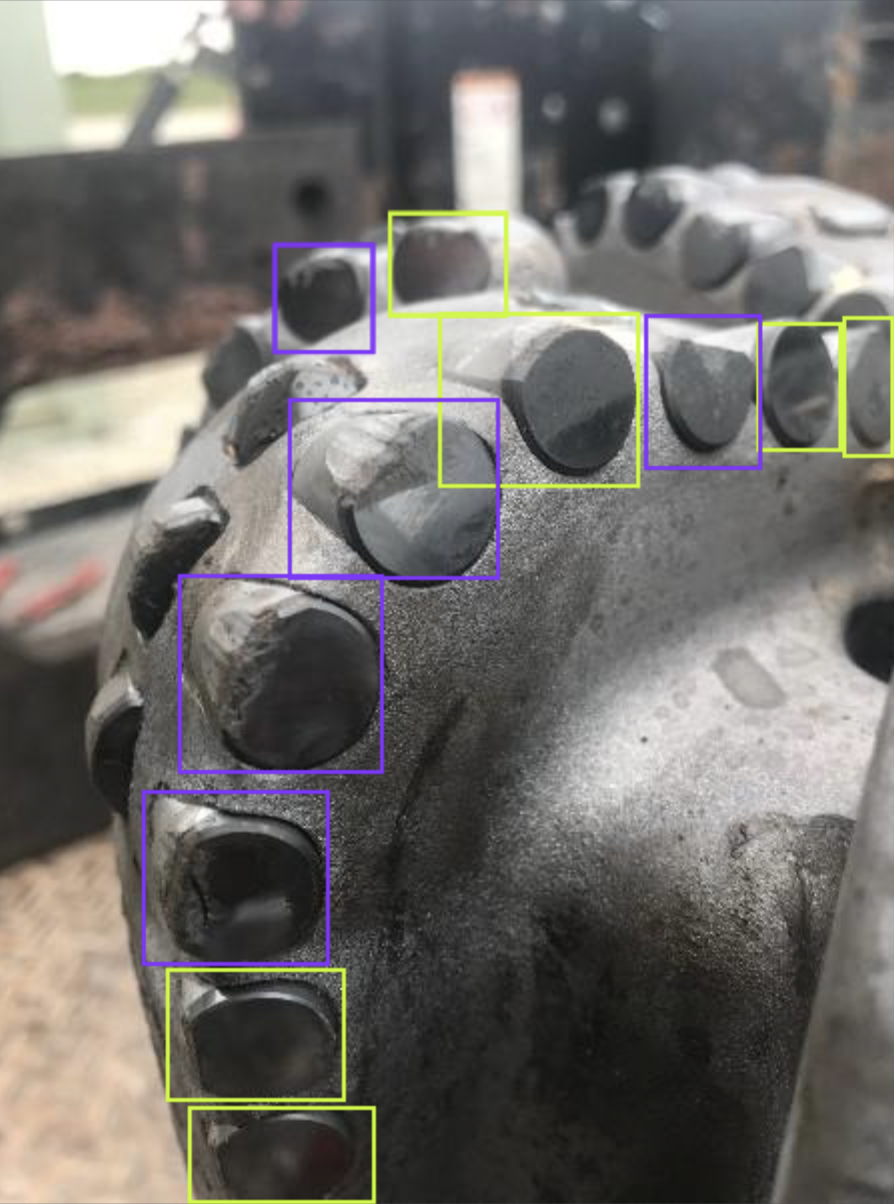}
  \caption{Left: Annotating only the location of cutters on the main blade, Right: Annotating the damage of cutters clearly visible.}
  \label{tab:meth-ann-0}
\end{figure}

\textit{Cutter Location Annotation.} 
This annotation falls under the object detection annotation category by drawing bounding boxes and labeling the location using the Roboflow web application~\cite{conf/icra/AlexandrovaTC15}, surrounding each cutter located on the main blade (facing forward) side, as illustrated in Fig.~\ref{tab:meth-ann-0}. We considered only the location of the cutters on the main blade so that the model could detect the foreground cutters without repetition. In this case, the model will map the background cutters to the negative sample to minimize false positives.

  

\textit{Cutter Type of Damage Annotation.} 
This type of annotation is more challenging for stratifying images by damage types. In this setup, and since location is irrelevant, most cutters present in the images will be annotated regardless of location. That is, except for cutters that are partially hidden or severally blurred, which may degrade your model's performance, we annotate all cutters with 75\% 

\textit{Identifying the main damages.} As each image may contain several types of damages, We propose the following annotation policy for identifying the main damages per location on the bit. The policy assigns damage importance order based on how much each damage contributes to identifying unique causes. The order is (1) Fractures, (2) Missing Cutters (or partially missing) due to fracture, delimitation, Ringouts, (3) Thermal Wear, and (4) Smooth Wear. Then, the procedure of annotating the failure cause is as follows: (1) Identify the damages on each side of the drill bit. (2) Per side, per location, identify the main damage following the damage importance order proposed. If there is a majority annotation with a lower rank, consider it the main annotation. (3) Per drill bit, per location, aggregate the main damages and identify the main damage following the damaged order proposed.

{\textit{Failure Cause Annotation Policy}} 
As there can be multiple failures caused by the same drill bit. We consider failure causes identification to be a multi-class, multi-label problem. Our annotation includes 8 classes for the 46 drill bits: Smooth Wear, Thermal Wear, stick-slip, Axial, Whirl, Hard Formation Transition, Soft Formation Transition, and Structural overload/nasal misplacement (under the label of coreout). This step is summarized in Table~\ref{tab:causes}. This is in addition to the green class when (almost) no damage is observed.

\begin{table}[h!]
\caption{Bit failure cause per main damage and location}
\label{tab:causes}
\resizebox{\columnwidth}{!}{%
\begin{tabular} {p{0.2\columnwidth}p{0.2\columnwidth}p{0.3\columnwidth}p{0.3\columnwidth}p{0.3\columnwidth}} 
\toprule
                                         & \textbf{Core}       & \textbf{Nose}             & \textbf{Shoulder}         & \textbf{Gauge}                                          \\ \midrule
\textbf{Smooth Wear}                     & Smooth Wear         & Smooth Wear               & Smooth Wear               & Smooth Wear                                             \\
\textbf{Thermal Wear (Low/Medium)}       & Thermal Wear        & Thermal Wear              & Thermal Wear              & Thermal Wear                                            \\
\textbf{Tangential Fracture}             & stick-slip          & Hard Formation Transition & Whirl (lateral)           & Whirl (lateral) if no other damages then inside casing \\
\textbf{Normal Fracture}                 & stick-slip          & Axial                     & Whirl (lateral)           & Whirl (lateral) if no other damages then inside casing \\
\textbf{Nose Ringout (High missing)}     &                     & Hard Formation Transition &                           &                                                         \\
\textbf{Shoulder Ringout (High missing)} &                     &                           & Soft formation transition &                                                         \\
\textbf{Core out (High missing)}         & Structural overload &                           &                           &                                                         \\
\textbf{Low Missing}                     & stick-slip          & Axial                     & Whirl (lateral)           & Whirl (lateral)                                         \\ \bottomrule
\end{tabular}%
}
\vspace{-10pt}\end{table}

\subsubsection{Preparing data for training} 
For the cutter damage and location detection models, we split the data into 36 drill bits image sets for training and 10 drill bits image sets for testing (around 79\% for training, and 21\% for testing). This split tests the robustness of the models and their ability to generalize, as the types of drill bits and types of damages are not necessarily evenly distributed.

\subsubsection{Dataset Augmentation} 
In order to increase the training set for the cutter detection related models,  three augmented proceedures, using Roboflow web application~\cite{conf/icra/AlexandrovaTC15}, which are: (1) Flip (Horizontal), (2)90° Rotate (Clockwise, Counter-Clockwise) (3) Brightness(Between -25\% and +25\%). This increased the total number of training images from 205 to 615 images.
 

\subsection{Metrics}
As we have two types of models, two sets of metrics will be used for evaluation. One set for the object detection model and the other for the classification model. 

\textit{Object Detection} The main metric considered for object detection is the mean Average Precision (mAP). It is calculated as the overlap criterion, as the Intersection over Union (IoU) is greater with respect to the ground truth box. With a set of pixels of $A$ as the proposed pixels and $B$ as the ground truth pixels for a certain object, the IoU is:
\begin{equation}
    IoU(A,B) = \frac{A \cap B}{A \cup  B}
\end{equation}
Each object can have only one label as a true positive. All other labels will be considered false positives~\cite{Everingham2009}. Undetected objects based on ground truth are considered false negatives. The Average Precision (AP) is computed by averaging the precision values on the precision-recall curve. The mAP0.5 is the average AP across classes. 

\begin{equation}
    mAP0.5 = \frac{1}{n} \sum_{k=1}^{k=n} AP_{k}, 
\end{equation}
where n is the number of classes, and $AP_{k}$ is the average precision for a class.

\textit{Failure Cause Classification} For the failure-cause classification problem, we will use the common classification metrics: accuracy, precision, recall, and f1-score.

\subsection{Cutter Detection Models} 
The detection of the cutter, along with its location and damage, consists of three steps: cutter location detection, cutter damage detection, and location and damage alignment. 

\textit{Cutter Location Detection.} We trained a YOLOv5 model using the dataset annotated for cutter location, with the default hyper-parameters, to detect only the cutters on the main blade along with their locations. Multiple types of drill bits, regardless of the damage, were involved in the training with different angles due to the augmentation. The locations considered are the top, Core, Nose, Shoulder, and Gauge.

\begin{figure}[!ht]
  \centering
  \includegraphics[width=0.75\columnwidth]{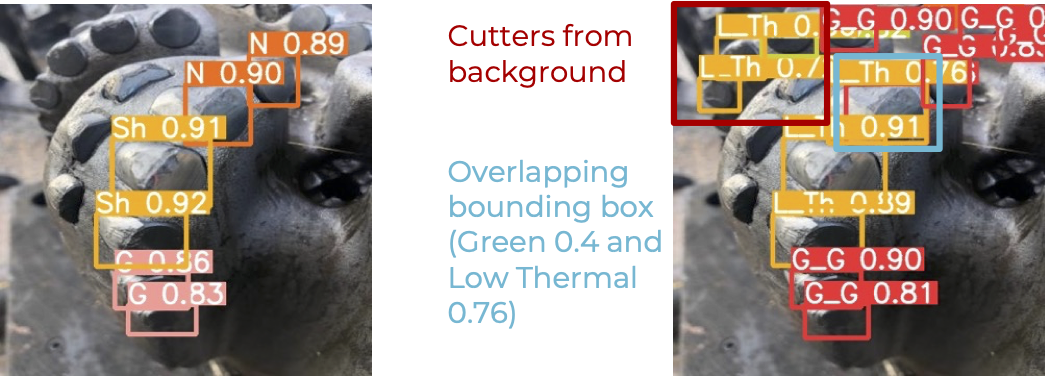}
  \caption{Multiple overlapping damages.}
  \label{tab:overlap}
\end{figure}

\textit{Cutter Damage Detection.} Similarly, we trained a YOLOv5 model using the dataset annotated for cutter damage type, with the default hyper-parameters, to detect the visible cutters along with their damages.  The damages considered are in Table~\ref{tab:cdmr} 
\textit{Location and Damage Alignment.} Since different models are used to detect the location and damage, we introduced an alignment step to detect the cutters and their damages along the main blade. In order to quantify the damage for each of the detected cutters on the main blade, we established the following strategy for each of the detected cutters from the cutter location detection model: (1) Calculate the Euclidean distance between the center of the bounding box and all the bounding boxes for the detected cutters on the cutter damage detection model. (2) Determine the overlapping damage annotated cutters identified by a Euclidean distance of less than 0.05 (this was determined empirically). (3) Check the confidence score to pick the overlapping object with the highest 0.5mA5 score, as in Figure~\ref{tab:overlap}.

\begin{algorithm}
\caption{RRFCI for unfractured cutters, \textbf{Input} : bitinfo which is a dictionary containing cutters' locations coupled with the damages}\label{alg:increasingsub}
\begin{algorithmic}[1]
\Procedure{isGreen}{}
   \If{The number of green cutters is higher than 80\% of the total bits detected}
      \State \Return True
   \EndIf
\EndProcedure
\Procedure{isSmoothWear}{}
   \If{The number of cutters with Smooth Wear is higher than the bits with Thermal Wear}
      \State \Return True
   \EndIf
\EndProcedure
\Procedure{isThermalWear}{}
   \If{The number of cutters with Thermal Wear on the Shoulders is higher than the number of main blades (7 in the bits studied)}
      \State \Return True
   \EndIf
   \If{The number of cutters with Thermal Wear at any location is higher than the number of main blades * 2 (14 in the bits studied)}
      \State \Return True
   \EndIf
\EndProcedure
\end{algorithmic}
\end{algorithm}

\begin{algorithm}
\caption{RRFCI for Ringouts, \textbf{Input} : bitinfo which is a dictionary containing cutters' locations coupled with the damages}\label{alg:increasingsub}
\begin{algorithmic}[1]

  \Procedure{isRingout}{} 
  \Comment{From the top view}
    \If{The top view has a Ringout 'RO'}
      \State \Return True
    \EndIf
  \EndProcedure

  \Procedure{isCoreout}{}
    \If{The number of missing core cutters is greater than 1}
      \State \Return True
    \EndIf
  \EndProcedure

  \Procedure{isNoseRingout}{}
    \If{The number of missing Nose cutters is greater than 5}
      \State \Return True
    \EndIf
    \If{The number of detected unmissing cutters is less than 10}
      \State \Return True
    \EndIf
    \If{'Shoulder\_RO' damage is detected for 'Shoulder\_RO' location}
      \State \Return True
    \EndIf
    \If{isRingout() and the number of missing Nose cutters is greater than the number of missing Shoulder cutters and not isCoreout() and the number of heavily damaged Nose cutters is greater than 2}
      \State \Return True
    \EndIf
  \EndProcedure

  \Procedure{isShoulderRingout}{}
    \If{The number of missing Shoulder cutters is greater than 5}
      \State \Return True
    \EndIf
    \If{'Shoulder\_RO' damage is detected for 'Shoulder\_RO' location}
      \State \Return True
    \EndIf
    \If{isRingout() and the number of missing Shoulder cutters is greater than the number of missing Nose cutters and not isCoreout() and the number of heavily damaged Shoulder cutters is greater than 2}
      \State \Return True
    \EndIf
  \EndProcedure
\end{algorithmic}
\end{algorithm}
 
\begin{algorithm}
\caption{RRFCI fractured cutters, \textbf{Input} : bitinfo which is a dictionary containing cutters' locations coupled with the damages}\label{alg:increasingsub}
\begin{algorithmic}[1]
\Procedure{isStickSlip}{}
  \If{ missing core cutters equals 2 and is not isNoseRingout() and is not isCoreout()} \Return True
    \EndIf
    \If{ heavily damage core cutters or medium Thermal Wear equals 2 and is not isNoseRingout() and is not isCoreout()} \Return True
  \EndIf
\EndProcedure

\Procedure{isAxial}{}
  \If{Presence of cutters with 'H' annotation and is not isNoseRingout() and is not isCoreout()} \Return True
    \EndIf
\If{Presence of cutters with Normal Fracture} \Return True
    \EndIf

    \If{Presence of cutters with medium Thermal Wear on Nose and Shoulder is 75\% green} \Return True
    \EndIf
        \If{Number of cutters with medium Thermal Wear on the Nose is greater than  1.5 * cutters with medium Thermal Wear on Shoulder} \Return True 
  \EndIf
\EndProcedure

\Procedure{isWhirl}{}
  \If{presence of heavily damaged on Gauge or (Shoulder and  not isShoulderRingout()) or the presence of Tangential Fracture on Nose}  \Return True
  \EndIf
\EndProcedure
\end{algorithmic}
\end{algorithm}

\subsection{Failure Cause Identification}
\textit{First Approach: DT and RF models.} We built DT and RF models using the default hyperparameters for the multi-class, multi-label classification problem, using the annotated failure cause dataset with 46 data points.  

\textit{Second Approach: Rule Based} In this approach, we designed a set of rules based on the cutters' damages and locations. These rules are based on Table~\ref{tab:causes}. This rule-based approach eliminates the need for annotation and the aggregation step required to adopt ML models. It also takes into account the propagating errors from the damage detection model. The pseudocode for the Rule-Based Failure Cause Identification (RRFCI) is in algorithms 
1-3. 

\subsection{Evaluation approach for the complete pipeline.}
To evaluate the complete pipeline, including the cutter location and damage detection models, the alignment, and then the ML and rule-based approaches for failure cause detection, we held out 10 bits randomly, numbered 15, 20, 32, 39, 47, 49, 50, 52, 57, and 58. These bits were not used in training the cutter detection models or the ML models in the final fitting to ensure a fair evaluation.

\section{Results and Analysis}

\subsection{Annotation Results.} 
\begin{figure}[h!]
  \centering
  \includegraphics[width=0.9\columnwidth]{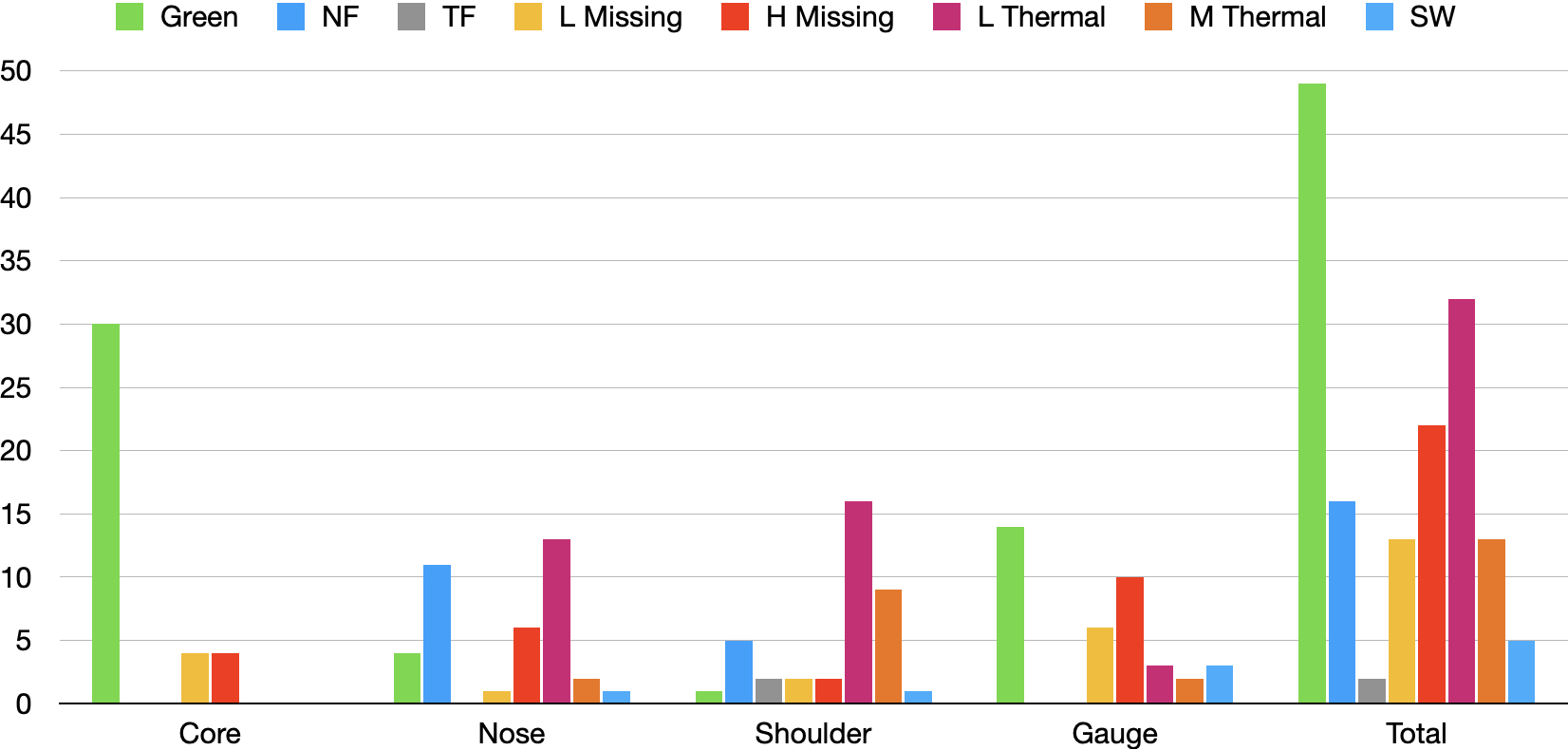}
  \caption{Dominant cause per location annotation.}
  \label{tab:res-ann-cause-1}
\end{figure}

The annotation results are evaluated on four levels: cutter location annotation, cutter damage annotation, identifying main damages, and drill bit failure cause. For the \textit{Cutter Location Annotation}, we annotated a total of 1909 cutters distributed over 295 drill bit images along with their locations on the main drill bits labels. Statistics of the annotation are present in Table~\ref{tab:res-ann-loc}. As for \textit{Cutter Damage Annotation}, we annotated a total of 3200 cutters distributed over 376 drill bit images along with their damage types as labels; summary of the annotations is present in Table~\ref{tab:res-ann-dmg2}. When it comes to \textit{Identifying the main damages}, the distribution is in Fig.~\ref{tab:res-ann-cause-1}, showing low Tangential Fracture and Smooth Wear representation overall.As for \textit{Failure Cause Annotation,}, the distribution of failure cause annotation indicates a lack of representation of stick-slips. Therefore, they it not be considered during evaluation. There is also an under-representation for Shoulder Ringout. Failure cause annotation distribution is illustrated in Table~\ref{tab:res-ann-cause-2}. 
\label{res}

\begin{table}[]
  \caption{Results of failure cause annotation.}
  \label{tab:res-ann-cause-2}
  \resizebox{\columnwidth}{!}{%
\begin{tabular}{@{}lllllllll@{}}
\toprule
Type of failure & Smooth Wear & Thermal Wear & Core out & Nose Ringout & Shoulder Ringout & stick-slip & Axial & Whirl \\ \midrule
count           & 10          & 39           & 9        & 12           & 3                & 0          & 13    & 29    \\ \bottomrule
\end{tabular}%
}
\vspace{-10pt}\end{table}

\begin{table}[h!]
\centering
\caption{Results of cutter location annotation.}
\label{tab:res-ann-loc}
\resizebox{0.5\columnwidth}{!}{%
\begin{tabular}{@{}lll@{}}
\toprule
\textbf{Location}   & \textbf{Class}     & \textbf{Annotations} \\ \midrule
Core                        & C                  & 381                  \\
Nose                        & N                  & 626                  \\
Shoulder                    & Sh                 & 647                  \\
Gauge                       & G                  & 647                  \\
Top                         & top                & 46                   \\
Shoulder RO                         & Shoulder RO                & 14                   \\

\textbf{Total (376 images)} & \textbf{6 classes} & \textbf{2361}        \\ \bottomrule
\end{tabular}
}
\end{table}

\begin{table}[h!]

\center
\caption{Results of cutter damage annotation.}
\label{tab:res-ann-dmg2}
\resizebox{\columnwidth}{!}{%
\begin{tabular}{@{}llll@{}}
\toprule
\textbf{Type of Damage}              & \textbf{Class}      & \textbf{Annotations} & \textbf{Comments} \\ \midrule
Green                                & G\_G                & 1,465                 & Over represented  \\
Low Thermal Wear                     & L\_Th               & 705                  & Over represented  \\
Medium Thermal Wear                  & M\_Th               & 256                  &                   \\
High (Delimitation, missing cutters) & H                   & 296                  &                   \\
Smooth Wear                          & L\_SW               & 320                  &                   \\
Normal Fracture                      & NF                  & 89                   & Under represented \\
No Ringout from top view             & no\_ro              & 26                   & Under represented \\
Ringout from top view                & RO                  & 19                   & Under represented \\
Nose and Shoulder missing area       & Shoulder\_RO        & 13                    & Under represented \\
Tangential Fracture                  & TF                  & 6                    & Under represented \\
Green with TF line present           & G\_TF               & 5                    & Under represented \\
\textbf{Total (367 images)}          & \textbf{11 classes} & \textbf{3200}        & \textbf{}         \\ \bottomrule
\end{tabular}%
}
\vspace{-5pt}\end{table}

\vspace{-\baselineskip}
\subsection{Cutter Detection Models Results.} The cutter detection results consists of 3 parts: detecting the location, damage, and the alignment.

\textit{Cutter Location model results.} The cutter location model achieved an mAP@.5 of 0.97 across the test set. Investigating, the test set images, we noticed that in most cases,, the model didn't identify cutters located on background blades, as designed. In rare cases, there was an overlap in labeling with vastly different confidence scores, which contributes to the object detection error. Details of the results with respect to some other metrics are shown in Table \ref{tab:clmr}, and samples of the testing results are in Fig. \ref{fig:sidl}. 


\begin{table}[]
\centering
\caption{Cutter location model results.}
\label{tab:clmr}
\resizebox{0.6\columnwidth}{!}{%
\begin{tabular}{@{}lllllll@{}}
\toprule
\textbf{Class}  & \textbf{Labels} & \textbf{P} & \textbf{R} & \textbf{mAP@.5} & \textbf{mAP@.5:.95:} \\ \midrule
\textbf{All}                & 515             & 0.957      & 0.95       & 0.973           & 0.699                \\
\textbf{C}                  & 84              & 0.949      & 0.917      & 0.955           & 0.641                \\
\textbf{G}                  & 140             & 0.973      & 0.943      & 0.971           & 0.658                \\
\textbf{N}                  & 139             & 0.954      & 0.914      & 0.961           & 0.647                \\
\textbf{Sh}                 & 142             & 0.943      & 0.979      & 0.984           & 0.725                \\
\textbf{Top}                & 10              & 0.966      & 1          & 0.995           & 0.826                \\ \bottomrule
\end{tabular}%
}
\vspace{-10pt}\end{table}

\begin{figure}[h!]
  \centering
  \includegraphics[width=0.8\columnwidth]{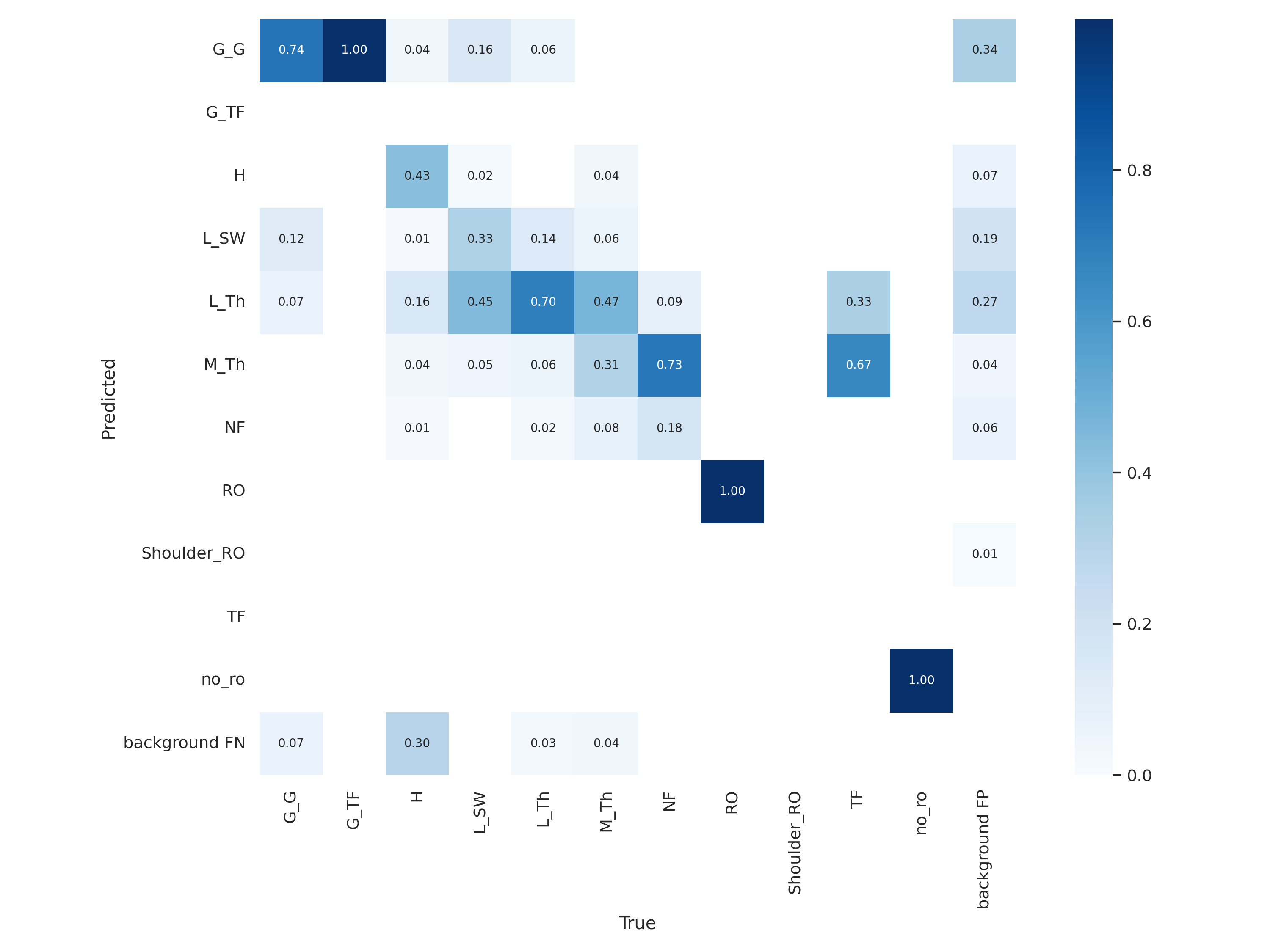}

  \caption{Confusion matrix of the cutter damage detection model.}
  \label{tab:conf-dmg}
\end{figure}

\begin{figure}
    \centering
    \includegraphics[width=0.3\columnwidth]{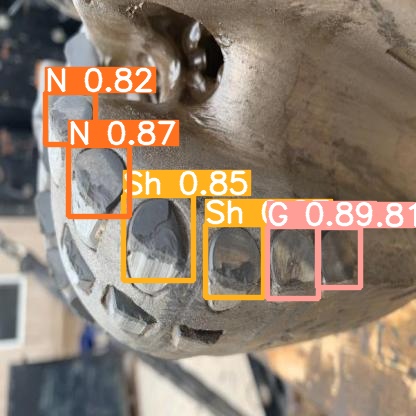} 
    \includegraphics[width=0.3\columnwidth]{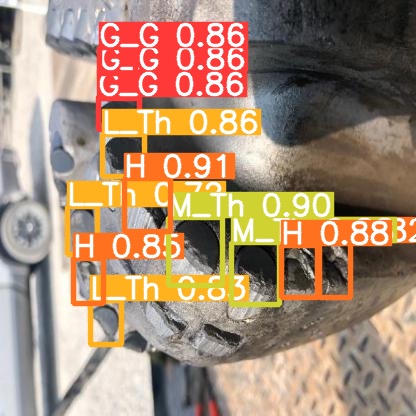}

    \caption{Samples of the damage location (left) and type detection results (right).}
    \label{fig:sidl}
\end{figure}

\textit{Cutter Damage model results.} The results of the damage model at this condition show a mAP@.5 of 0.49, shown in Table \ref{tab:cdmr}. The main reason for the high error is due to the model inability to identify two classes (TF,G\_TF). Removing these two classes improved the mAP to mAP@0.6 Investigating the resulting predictions, where examples are shown in Figure \ref{fig:sidl}, and the confusion matrix is depicted in Figure \ref{tab:conf-dmg}, we identify the following reason behind this performance:
\begin{enumerate}
    \item Missing annotation: as mentioned earlier, the annotation for the background cutters' damage is not a clear cut annotation. Therefore, the model detected many unlabeled cutters with high confidence, affecting accuracy. This is noted from the images and the confusion matrix, where many cutters are detected with green and low thermal types of damage. 
    \item Overlapping predicted bounding boxes: a problem degrading performance, especially for visually similar damages. This can be accounted for by considering the confidence score.
    \item Normal Fracture is confused with medium Thermal Wear, as the spalling in both cases is sometimes extremely similar when excluding the wear, which is not obvious depending on the angle. However, it is not as severe as in the confusion matrix, as many results from overlap labeling without considering the confidence score that usually favors the correct damage type.
    \item Smooth Wear is, in many cases, confused with low Thermal Wear. This is expected if there is a fine line between them during the annotation.
    \item Tangential Fracture (TF) resulting from Whirl was not detected and confused by Thermal Wear (medium or low) depending of the severity of the damage. This is expected as TF resulting from Whirl has only 3 labels (cutter) in the training data. As a note, we only label TF for TF resulting from Whirl. When TF is due to structural overload, it is most likely among the cutters labeled with (H).
\end{enumerate}

\begin{table}[]
\centering
\caption{Cutter Damage model results.}
\label{tab:cdmr}
\resizebox{0.9\columnwidth}{!}{%

\begin{tabular}{@{}lllllll@{}}
\toprule
\textbf{Class}  & \textbf{Labels} & \textbf{P} & \textbf{R} & \textbf{mAP@.5} & \textbf{mAP@.5:.95:} \\ \midrule
\textbf{all}                 & 731             & 0.646      & 0.532      & 0.49                   & 0.35                 \\
\textbf{Green (G\_G)}                & 349             & 0.77       & 0.777      & 0.851                  & 0.534                \\
\textbf{Green with TF line (G\_TF)}               & 1               & 1          & 0          & 0.00318                & 0.00254              \\
\textbf{Missing (H)}                   & 70              & 0.628      & 0.443      & 0.505                  & 0.282                \\
\textbf{Smooth Wear (L\_SW)}               & 58              & 0.218      & 0.481      & 0.29                   & 0.225                \\
\textbf{Low Thermal Wear(L\_Th)}               & 178             & 0.516      & 0.753      & 0.619                  & 0.444                \\
\textbf{Medium Thermal Wear (M\_Th)}               & 51              & 0.404      & 0.412      & 0.4                    & 0.278                \\
\textbf{Normal Fracture (NF)}                  & 11              & 0.202      & 0.455      & 0.215                  & 0.134                \\
\textbf{Ringout (RO)}                  & 4               & 0.73       & 1          & 0.995                  & 0.696                \\
\textbf{Tangential Fracture (TF)}                  & 3               & 1          & 0          & 0.0243                 & 0.0206               \\
\textbf{No Ringout (no\_ro)}              & 6               & 0.99       & 1          & 0.995                  & 0.885                \\ \bottomrule
\end{tabular}%
}
\vspace{-10pt}\end{table}

\textit{Location and damage alignment.}
The results of the alignment process are very promising, but a manual examination showed minor confusion regarding the damages.  

\subsection{Failure Cause Identification Results.} 

Looking at the performance of the ML models in Table \ref{tab:RF}, we notice the high accuracy with six classes above 80\%, and the remaining class (Whirl) is at 59\% and 72\%  for the DT and RF models, respectively. However, as the classes are highly imbalanced, as shown in Fig. \ref{tab:res-ann-cause-2}, the F1-score gives a better performance estimate in highly imbalanced cases. The macro-average F1-score is 0.78 and 0.79 for the DT and RF models, respectively. Both models were not able to detect True Positive instances of Hard to Soft formation transition failure caused and repetitively mistaken Smooth Wear with Thermal Wear. On the other hand, the rule-based failure causes detection approach accuracy, also in Table \ref{tab:RF}, reached 96\% and had a macro-average F1-score of 0.94. Surprisingly, this outperforms both ML models. The rule-based approach was able to identify Hard to Soft formation transition failure cause perfectly (1.0 F-score). Although it did mistake Smooth Wear with Thermal Wear in some cases, the rule-based model achieved an F1-score of 0.84 on this type of cause, outperforming the ML model.

\subsection{Evaluation of the Complete Pipeline (YOLO Bit Forensics).} 
The results of the complete pipeline evaluation are in Table~\ref{fnal}. Although the two ML models had a similar performance during 
the leave-one-out evaluation strategy, we notice that the RF model was able to outperform the DT model as it identified 75\% of the total causes in comparison to 62.5\% for the DT model. 
The pipeline achieved the best when deploying the rule-based approach instead of the ML approach. Under this setup, it identified 100\% of the failure causes. It falsely identified one extra failure case for 2 bits, one related to Whirl and the other to Axial, which are due to propagating errors from the damage detection model.
\begin{table}[]
\centering
\caption{Results of Bit Failure Cause identification approaches.}
\label{tab:RF}
\resizebox{\columnwidth}{!}{%
\begin{tabular}{p{0.12\columnwidth} p{0.15\columnwidth} p{0.16\columnwidth} p{0.1\columnwidth} p{0.2\columnwidth} p{0.22\columnwidth} p{0.12\columnwidth} p{0.12\columnwidth} p{0.15\columnwidth}}%
\toprule
\textbf{}          & \textbf{Smooth Wear} & \textbf{Thermal Wear} & \textbf{Core out} & \textbf{Hard Formation transition} & \textbf{Soft Formation transition} & \textbf{Axial} & \textbf{Whirl} & \textbf{macro-Average} \\ \midrule

\multicolumn{9}{c}{\textbf{Decision Tree}} \\ \midrule
\textbf{Accuracy}  & 82.61\%              & 86.96\%               & 89.13\%           & 97.83\%                            & 89.13\%                            & 97.83\%        & 58.70\%        & 86.02\%                \\
\textbf{Precision} & 0.83                 & 0.95                  & 0.80              & 1.00                               & 0.00                               & 0.93           & 0.67           & 86.33\%                \\
\textbf{Recall}    & 0.42                 & 0.91                  & 0.50              & 0.92                               & 0.00                               & 1.00           & 0.69           & 73.94\%                \\
\textbf{F1 score}  & 0.56                 & 0.93                  & 0.62              & 0.96                               & -                                  & 0.96           & 0.68           & 78.34\%                \\ \midrule

\multicolumn{9}{c}{\textbf{Random Forest}} \\ \midrule
\textbf{Accuracy}  & 80.43\%              & 91.30\%               & 93.48\%           & 97.83\%                            & 91.30\%                            & 93.48\%        & 71.74\%        & 88.51\%                \\
\textbf{Precision} & 0.75                 & 0.95                  & 1.00              & 1.00                               & 0.00                               & 0.92           & 0.81           & 90.58\%                \\
\textbf{Recall}    & 0.27                 & 0.95                  & 0.63              & 0.92                               & 0.00                               & 0.85           & 0.73           & 72.56\%                \\
\textbf{F1 score}  & 0.40                 & 0.95                  & 0.77              & 0.96                               & -                                  & 0.88           & 0.77           & 78.91\%                \\ \midrule

\multicolumn{9}{c}{\textbf{Rule-based approach}} \\ \midrule
\textbf{Accuracy}  & 93.48\%              & 93.48\%               & 97.83\%           & 100.00\%                           & 100.00\%                           & 93.48\%        & 93.48\%        & 95.96\%                \\
\textbf{Precision} & 0.89                 & 0.97                  & 1.00              & 1.00                               & 1.00                               & 0.81           & 0.91           & 0.94                   \\
\textbf{Recall}    & 0.80                 & 0.95                  & 0.89              & 1.00                               & 1.00                               & 1.00           & 1.00           & 0.95                   \\
\textbf{F1 score}  & 0.84                 & 0.96                  & 0.94              & 1.00                               & 1.00                               & 0.90           & 0.96           & 0.94                   \\ \bottomrule
\end{tabular}%
}
\end{table}

\begin{table}[]
\centering
\caption{Results of complete pipeline. }
\label{fnal}
\resizebox{\columnwidth}{!}{%
\begin{tabular} {p{0.05\columnwidth}p{0.25\columnwidth}p{0.25\columnwidth}p{0.25\columnwidth}p{0.25\columnwidth}} 

\toprule
\textbf{Bit} & \textbf{Existing Failure Causes}                              & \textbf{Detected Failure Causes by DT}                    & \textbf{Detected Failure Causes by RF}                                                               & \textbf{Detected Failure Causes by rule-based approach}                                              \\ \midrule
\textbf{15}  &  \textbf{Thermal Wear, Soft to Hard Formation Transition}               & Thermal Wear, Soft to Hard Formation Transition, Whirl    & Thermal Wear, Soft to Hard Formation Transition, Whirl                                               &  \textbf{Thermal Wear, Soft to Hard Formation Transition}                                                      \\
\textbf{20}  &  \textbf{Thermal Wear}                                                 & Smooth Wear                                               & Smooth Wear                                                                                          &  \textbf{Thermal Wear}                                                                                         \\
\textbf{32}  &  \textbf{Thermal Wear, Soft to Hard Formation Transition}               & Thermal Wear, Soft to Hard Formation Transition, Whirl    & Thermal Wear, Soft to Hard Formation Transition, Whirl                                               &  Thermal Wear, Soft to Hard Formation Transition, Whirl                                               \\
\textbf{39}  &  \textbf{Thermal Wear, Axial, Whirl}                                    & Smooth Wear, Whirl                                        & Thermal Wear, Axial                                                                                  &  \textbf{Thermal Wear, Axial, Whirl}                                                                           \\
\textbf{47}  &  \textbf{Thermal Wear, Whirl}                                           & Thermal Wear                                              & Thermal Wear                                                                                         & Thermal Wear, Axial, Whirl                                                                           \\
\textbf{49}  &  \textbf{Thermal Wear, Soft to Hard Formation Transition, Whirl}        & Thermal Wear, Soft to Hard Formation Transition           & Thermal Wear,  Soft to Hard Formation Transition                                                     &  \textbf{Thermal Wear, Soft to Hard Formation Transition, Whirl}                                               \\
\textbf{50}  &  \textbf{Smooth Wear, Whirl}                                            & Thermal Wear, Whirl                                       & Thermal Wear, Whirl                                                                                  &  \textbf{Smooth Wear, Whirl}                                                                                   \\
\textbf{52}  &  \textbf{Thermal Wear, Whirl}                                           & Thermal Wear, Smooth Wear, Whirl                          & Thermal Wear, Smooth Wear, Whirl                                                                     &  \textbf{Thermal Wear, Whirl}                                                                                  \\
\textbf{57}  &  \textbf{Thermal Wear, Soft to Hard Formation Transition, Axial, Whirl} & Thermal Wear, Smooth Wear, Axial                          &  \textbf{Thermal Wear, Soft to Hard Formation Transition, Axial, Whirl}                                       &   \textbf{Thermal Wear, Soft to Hard Formation Transition, Axial, Whirl}                                        \\
\textbf{58}  &  \textbf{Thermal Wear, Axial, Whirl}                                    & Thermal Wear, Smooth Wear, Whirl                          & Thermal Wear, Smooth Wear, Whirl                                                                     &  \textbf{Thermal Wear, Axial, Whirl}                                                                           \\ \midrule
 \textbf{Total}        &  \textbf{24 failure causes}                                             & 15 correctly detected Failure Causes and 8 False detected & 18 correctly detected Failure Causes  and 6 False detected &  \textbf{24 correctly detected Failure Causes and 2 False detected}
\end{tabular}%
}
\vspace{-10pt}\end{table}

\section{Discussion}
\label{des}
Our evaluation study has shown that our rule-based approach outperformed our ML approach for detecting failure causes. This can be attributed to the following points: (1) During the feature engineering process to extract the features to build the ML models on, the hierarchy proposed to identify the main cause of damage per location over simplified the problem as some causes were suppressed due to having a lower profile or appearing only in one or two cutters. This could be mitigated in the future by using the full damage profile. (2) As the ML models were fit based on the ground truth annotation of the cutter damages, it cannot deal with errors propagating from the damage detection model. This could be mitigated by data augmentation through adding some perturbation or noise to the features. (3) When a lack of balanced representative data exists, the learning of ML models becomes limited, and therefore, an expert-driven rule-based model performs better. This could be mitigated by increasing the instances of the underrepresented classes in the training data.

\section{Conclusion}
\label{conc}
Identifying the root cause of drill bit failure rapidly aids in preparation for future bit runs and prolongs the life expectancy of the drill bit. Performing forensics is subjective and relies on human expertise, which is expensive and time-consuming. Our proposed approach aims to automate bit damage identification and classification regarding location, type, and root cause using a YOLOv5 model and a rule-based approach to identify the associated type and root cause of drill bit failure. The evaluation results provide good evidence that (1) our models were highly accurate in detecting the location of the bit and the type of damage with 0.97 mAP and 0.66-0.49 mAP, respectively. (2) The rule-based approach to classifying the failure cause achieved an accuracy of 95\% and a macro-average F1-score of 0.94. (3) The complete pipeline was able to identify all of the 24 failure causes present in the test set. In the future, we aim to enhance the models by obtaining a diverse, larger dataset of bit images. Furthermore, we will investigate the performance of our pipeline in the case of images suffering from low-quality and noise.

\section*{Acknowledgment}
The authors acknowledge the support of King Fahd University of Petroleum and Minerals in the development of this work. They would like to thank the SPE DUPTS 2022 Competition team and especially Dr.Pradeepkumar Ashok for providing the data and guidance. Thanks are extended to Meshal Al-Otaibi and Jawhara Al-Mulhim for the knowledge shared and support.

\bibliographystyle{IEEEtran}
\bibliography{conference_101719}

\end{document}